\documentclass[11pt,a4paper,table]{article}
\usepackage[hyperref]{eacl2021}
\usepackage{times}
\usepackage{latexsym}

\usepackage{microtype}
\usepackage[T1]{fontenc}
\usepackage{graphicx}
\usepackage{hyperref}
\usepackage{longtable}
\usepackage{bm}
\usepackage{amsmath}
\usepackage{amssymb}
\usepackage{multirow}
\usepackage{xcolor}
\usepackage{placeins}

\aclfinalcopy

\newcommand{\eg}{\textit{e.g.} }
\newcommand{\ie}{\textit{i.e.} }

\newcommand{\tacred}{TACRED}
\newcommand{\docred}{DocRED}
\newcommand{\roberta}{RoBERTa}
\newcommand{\distant}{Distant Supervision} 

\newcommand{\spike}{SPIKE}
\newcommand{\syntacticby}{Syntactic Search by Example}

\title{Bootstrapping Relation Extractors \\using Syntactic Search by Examples}

\author{Matan Eyal\textsuperscript{1} \quad Asaf Amrami\textsuperscript{1, 2} \quad Hillel Taub-Tabib\textsuperscript{1} \quad Yoav Goldberg\textsuperscript{1, 2}  \\
  \textsuperscript{1}Allen Institute for AI, Tel Aviv, Israel \\
  \textsuperscript{2}Bar Ilan University, Ramat-Gan, Israel \\
  \texttt{{matane,asafa,hillelt,yoavg}@allenai.org}
  }

\date{}

\begin{document}
\maketitle
\begin{abstract}

The advent of neural-networks in NLP brought with it substantial improvements in supervised relation extraction. However, obtaining a sufficient quantity of training data remains a key challenge.
In this work we propose a process for bootstrapping training datasets which can be performed quickly by non-NLP-experts. We take advantage of search engines over syntactic-graphs (Such as \citet{syntactic-search-2020}) which expose a friendly by-example syntax. We use these to obtain positive examples by searching for sentences that are syntactically similar to user input examples.
We apply this technique to relations from \tacred~and \docred~and show that the resulting models are competitive with models trained on manually annotated data and on data obtained from distant supervision. The models also outperform models trained using NLG data augmentation techniques.
Extending the search-based approach with the NLG method further improves the results.
\end{abstract}

\section{Introduction}
\label{sec:introduction}
The goal of Relation Extraction (RE) is to find and classify instances of certain relations in raw text. We denote a binary relation instance, \ie a relation instance with two arguments, with a tuple $\bm{x} = (\bm{s}, e^1, e^2, r)$, where $\bm{s} = [w_0 \cdots w_n]$ is a sequence of sentence tokens, $e^1, e^2$ are entity mentions within $\bm{s}$ corresponding to the first and second relation argument, respectively, and $r \in R \cup \{\varnothing\}$ is a relation label from a set of predefined relations of interest, or an indication of `no-relation'. In binary classification our goal is to classify whether, according to $\bm{s}$, the entity mentions, $e^1$ and $e^2$, satisfy $r$, the relation label. For such classification we require a training dataset $X$, comprised of $X_p$, a set of positive examples, representing the relation of interest, and  $X_n$, a set of negatives examples. 

The success of recent papers \cite{soares2019matching, murty2020expbert} in supervised RE is fueled by advances in deep learning, but also, crucially, by the availability of a large training set such as TACRED \cite{zhang2017tacred}, containing tens of thousands of training examples.
For most relations of interest, such training data is not available.

\textbf{In this work we examine methods to inexpensively construct $X_p$ and $X_n$, in cases where a training set is not available. We are especially interested in constructing the positive set, $X_p$.}

In contrast to common NLP tasks like POS tagging, entity extraction and dependency parsing, the task of relation extraction exhibits a much larger degree of label sparsity. For some relations, even when considering only sentences with entities of the relevant types, the ratio between positive and negative examples is highly skewed toward the latter and obtaining a modest amount of positive examples will require a laborious annotation effort (see \S\ref{data-annotation-challenge}). While manual annotation of large datasets is a viable approach, it typically requires contracting a team of professional annotators \cite{ace, kbp} or crowd workers \cite{zhang2017tacred, yao2019docred} and is not well suited for smaller projects or for ad-hoc extraction tasks.

Our main contribution in this paper is a new methodology built on top of \citet{syntactic-search-2020} for cheaply obtaining large datasets (\S\ref{method:syntactic}).  \citet{syntactic-search-2020} proposed a syntactic search engine that given a lightly annotated example sentence, retrieves new sentences with a similar syntactic structure from a pre-annotated dataset. 
Our syntactic search bootstrapping method requires a small number of manually curated positive example sentences. Then the search engine matches are used as training data for ML models. We evaluate this approach comparing to human annotated data of varying sizes.

While this method shows promising results with very few user input examples, we also test the impact on performance when more examples are used. One technique for obtaining an abundance of examples uses recent Natural Language Generation (NLG) models (\S\ref{method:search-and-generation}). It has been shown in recent papers \cite{wei2019eda, anaby2019not, kumar2020data, amin2020exploring, russo2020control} that generating abundance of training examples can improve classifier performance. We aim to check whether this can improve our syntactic search method as well.


We evaluate the proposed methodologies by training DL classifiers on the obtained data. 

\paragraph{We show that:}

(1) Syntactic patterns are competitive at bootstrapping training data for ML, even with as little as 3 patterns;\\ 
(2) Training DL models over the output of syntactic patterns can significantly improve both recall and F1 over a rule based approach which uses the patterns directly;\\
(3) Training ML models over the output of syntactic patterns performs better than training models over recently popular NLG data augmentation techniques;\\
(4) Augmenting the output of syntactic patterns using NLG techniques is often helpful;\\
(5) Different relations benefit from different strategies.

The code for all our experiments alongside the generation outputs is publicly available\footnote{ \href{https://www.github.com/mataney/BootstrappingRelationExtractors}{github.com/mataney/BootstrappingRelationExtractors}}.
\section{Related Work}

\noindent\textbf{Distant Supervision.} Since its introduction, Distant Supervision \cite{mintz2009distant} has established itself as a viable alternative to manual annotation. \distant~assumes the availability of a knowledge base (KB) of $\langle e^1, r, e^2 \rangle $ triplets where $e^1, e^2$ are entities known to satisfy relation $r$.
To obtain training examples for a relation $r$, we sample sentences from a large background corpus: sentences which include entity pairs listed in the KB as satisfying $r$ are labeled positive, the remaining sentences are labeled negative (potentially after satisfying additional constraints). While effective in some cases, the reliance on large pre-existing KBs is a significant limitation. Such KBs are not usually available and the cost of constructing them is high.

\noindent\textbf{Bootstrapping from Rules, Snorkel.} To eliminate the reliance on external KBs, \citet{angeli2015bootstrapped} used the predictions of a rule based extractor on a large corpus to train a first iteration of a statistical extractor. They then continued to refine the extractor through self-training. 

Another system which can optionally utilize rules instead of external KBs is Snorkel \cite{snorkel}. Snorkel is implementing the data-programming paradigm \cite{data-programming} where ML models are trained in three stages: (i) users write labeling functions that weakly label data points using arbitrary heuristics (\eg extraction rules); (ii) the system learns a re-weighted combination of the labeling functions by explicitly modeling the actual distribution of each class. The results are often precise but low-recall; and (iii)  The system uses discriminative models to increase recall while preserving precision. 

The techniques used by \citet{angeli2015bootstrapped} and Snorkel can be effective in increasing the accuracy of the initial labeling rules, but coming up with ``good enough" initial rules remains a major challenge. In this sense, the search-based methods suggested in this work for bootstrapping RE datasets are complimentary and can be plugged in as a first step in these multi-step solutions.

Only few papers can be directly compared to our paper and use matches as training-data for ML classifiers. One paper similar in that sense is \citet{angeli2013stanford} which claims that training a classifier using search-based examples works better than traditional bootstrapping methods. See \S\ref{sec:synt_search_results} for further compression with \citet{angeli2013stanford}.

\noindent\textbf{Augmentation Through Generation.} Similarly to our Example Generation approach, recent papers \cite{anaby2019not, kumar2020data} suggest using pre-trained language models for data augmentation. In both these papers, the authors suggest prepending class labels to generative models in order to augment the number of instances for classes with a small number of examples. In contrast to these papers we use language models in a zero-shot context, and rather than requiring existing labeled examples of the relevant relation, we propose to manually label the generated samples.

\section{The data annotation challenge}
\label{data-annotation-challenge}
In contrast to linguistic annotation tasks such as parts-of-speech, syntactic-trees or semantic roles, annotating data for relation-extraction does not require special expertise. Annotation can be easily performed by a motivated native speaker of the language (in case of "every-day" relations such as those available in TACRED and DocRED) or by a domain expert (in case of "specialized" relations such as in biomedicine or law). Annotating a given sentence for a given relation takes roughly the time it takes to read and understand the sentence. So what stops us from obtaining large amounts of annotated data for ML?

The annotation challenge lies in \emph{relation sparsity} in the wild.
In an attempt to get a perspective on this issue, let's consider the \emph{founded-by} relation between a PERSON and an ORG, as attested in the TACRED corpus. Assuming we consider only sentences that contain both a person mention and an organization mention, how many sentences do we have to annotate before we reach, for example, 10 positive examples? The TACRED training set has 124 \emph{founded-by} instances, as well as 6947 "negative" instances with matching entity types ("negative" examples are either other relations, or \emph{no-relation}). This 1-out-of-57 ratio indicates that we will likely sample 56 "negative" sentences before hitting a positive instance.\footnote{See Appendix \ref{appendix:effort} for similar distributions over all relations.} This ratio is overly optimistic, as the annotations in the TACRED corpus are already very skewed in favor of positive examples. Even under this very optimistic scenario, we will need to annotate 570 sentences to recover 10 positive examples. 
The cost of annotation, then, is not in annotating each individual positive sentence, but in finding the sentences to annotate in the first place. Therefore, we should seek for methods that point towards probable positive instances.

In this paper, we present two methods, the first returns close to 1-out-of-1 positive ratio, although with low syntactic diversity, and a second method with roughly 1-out-of-3 positive ratio.

\section{Problem Statement and Setup}
\label{sec:problem_statment}

We are interested in the problem of obtaining a relation classifier for a binary relation, when no a-priori annotated training data for this relation is available. We seek a methodology that will allow to create an effective extractor, using a minimal amount of data annotation effort.

We compare four approaches -- manual annotation, syntactic-search, manual annotation over generated examples, and a combination of the last two -- to be described in later sections. Here, we discuss setup which is shared to all experiments. 

In order to evaluate the methodology on multiple datasets with similar relations, we chose a set of relations that appear in both the \tacred~\cite{zhang2017tacred} and \docred~\cite{yao2019docred} datasets with at least 50 development examples\footnote{\textit{org:country of headquarters}, \textit{org:founded by}, \textit{per:children}, \textit{per:city of death}, \textit{per:date of death}, \textit{per:origin}, \textit{per:religion}, \textit{per:spouse} for \tacred, and similarly \textit{headquarters location}, \textit{founded by}, \textit{child}, \textit{place of death}, \textit{date of death},  \textit{country of origin}, \textit{religion}, \textit{spouse} for \docred.}.

To quantify the performance of our methodology we assess it comparing to varying amounts of manually annotated data. In our settings, large amounts of supervised examples represent upper bound for our bootstrapping methods and are not expected.

While relation extraction is often considered as a multi-class classification problem (``find the occurrences of any of these possible relations''), we instead treat the relations separately, training a binary classifier for each one. We believe this is more representative of a user who wishes to target a low number of relations, who is likely to conduct data collection and evaluation for one relation at a time.
\\[0.5em]\textbf{Obtaining Negative Examples}$\;$
When training a binary classifier, it is required to include a set of negative examples alongside the list of positive examples. In all our experiments we obtain negative examples by looking for sentences that contain entity types that are compatible with the relation (i.e, for the \emph{founded-by} relation we sample sentences that include both a PERSON and an ORG). In our syntactic based methods we sample from the same domain as our positive examples (Wikipedia) and then filter this list by removing sentences in which the entities are connected by a syntactic pattern which is attested by the positive examples. For the supervised baselines of various sizes, we obtain negative examples by sampling them from the annotated training set, without replacement.\footnote{
The \textbf{positive to negative} ratio in training data has an effect on the resulting model's quality.
We experimented with positive-to-negative ratios of 1, 5, 10 and 20, as well as with a ``match the dev-set'' ratio. We found a ratio of 10 negative examples for each positive sentence to performs well.}\\[0.5em]
\textbf{Datasets}$\;$
We used two datasets to explore our different methods. \tacred~\cite{zhang2017tacred}, a large-scale multi-class relation extraction dataset built over newswire and web text. And \docred~\cite{yao2019docred}, a dataset for document level RE, and similarly designed for multi-class prediction. Per our setup above, we changed the setting of both datasets to per relation binary classification. As our main goal in this paper is to evaluate different bootstrapping methods, and not novel methods for document-level relation extraction, we chose to include only instances with single supporting sentence in \docred~(\ie sentence level relations). As \docred's labelled test set is not publicly available, we used the development set as our test set and used 20\% of the train set as development set. 
\\[0.5em]\textbf{Models}$\;$
Our classifiers throughout the following experiments are based on the Entity Markers architecture \cite{soares2019matching}. In the paper, the authors proposed wrapping the relation arguments with marker tokens (\eg $[E1_{start}]$ John $[E1_{end}]$ was born in $[E2_{start}]$ 1948 $[E2_{end}]$). The altered text is then passed as input to a BERT model \cite{devlin2018bert} where the relation between the two entities is represented by the concatenation of the final hidden states corresponding to their respective start tokens. Finally, this representation is fed into a classification head and the model is fine-tuned for relation classification. \textit{cf.} \cite{soares2019matching} for more details. We use a similar model with the exception that we use a more recent pretrained language model, \roberta~\cite{liu2019roberta}, and perform binary, rather than multi-class, classification.

In all of the following experiments we trained our model with 3 different random seeds to lower variance introduced to the model with different initializations, and report the average score. 
At inference time we set the prediction threshold value for the test set to be the cut-off value that maximized F1 over the development set.

\vspace{-0.3em}
\section{Manual Annotation Baseline}
\noindent\textbf{Setup.}
Our comparison point throughout the paper is a model trained on traditionally-collected annotated data. 
We sample increasing-sized annotated sets from TACRED and DocRED, containing 55, 110, 220, 550, and 1100 examples. These correspond to 5, 10, 20, 50, 100 positive examples with 50, 100, 200, 500, 1,000 negatives examples. Additionally, we measure the performance on these datasets when using all available positive examples for each relation.

\noindent\textbf{Results}
Listed in the top rows of Table \ref{tab:f1_table}, averaged over all relations. Unsurprisingly, increasing the number of examples increases performance, with the exception of DocRED on which using all positive labels performs slightly worse than using 100 sampled positive examples for each relation, we attribute this to sampling noise.
\docred~scores are generally lower than TACRED scores. This is 
because of the way we constructed the development and test sets: while in \tacred's development set each sentence includes a single entity pair with a single relation, in \docred, we pass all possible sentences with entity pairs of the same type as the evaluated relation as possible candidates. This dramatically increases the number of candidates, and by that of possible type I errors. Moreover, as we included only examples with exactly one supporting sentence, the number of positive examples is low for some of the relations. All of this effects \docred~classification scores comparing to \tacred. 

Importantly, in all these experiments, the number of annotated examples used is significantly higher than the number used in our Syntactic Search experiments (3 examples in total).

\begin{figure}[t]
\centering
\includegraphics[width=\linewidth]{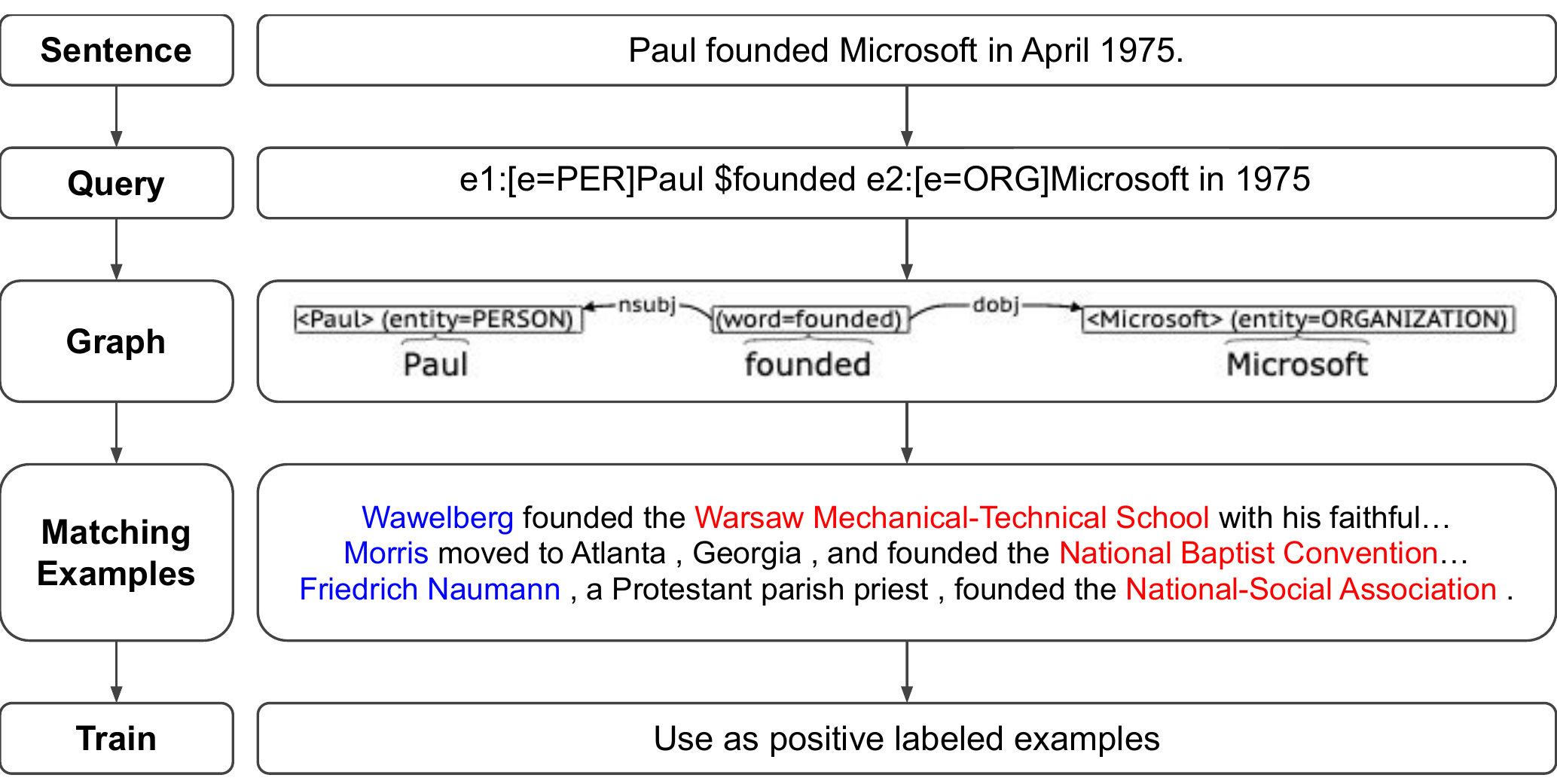}
\caption{Flow of the \syntacticby~method. For details, see \S\ref{method:syntactic}.} 
\label{fig:search-flow}
\end{figure}
\section{\syntacticby}
\label{method:syntactic}

We consider this section to be the main contribution of the work. We show that:\\
 (i) with modern DL modeling, effective relation extractors can be trained using sentences derived from less than a handful of syntactic patterns; and\\ (ii) through the use of by-example syntactic search engines, one can construct these patterns very quickly, without needing to understand syntax.

To explain the suggested workflow, let's consider a user who wants to train a relation extraction binary classifier for the \emph{founded by} relation, and has a single example sentence, ``Paul founded Microsoft in April 1975''. Patterns over syntactic structures and entity types are very effective for deriving high-precision extraction templates. For example, searching for sentences containing the word ``founded'' with an \emph{nsubj} dependency of type PERSON and \emph{dobj} dependency of type ORG, will return many matches for the \emph{founded-by} relations. 

There are two issues with this approach (1) while high-precision, the recall of the patterns is low; and (2) syntactic patterns require both linguistic and computational expertise to specify and execute.

The premise of this paper is that the low recall can be offset by machine learning. The sentences resulting from syntactic search over a few patterns are diverse enough that an ML model trained over them manages to generalize from the specific syntactic pattern and identify a broader range of cases, increasing recall substantially. We show this is indeed the case.

To overcome the need for linguistic expertise we propose using a by-example syntactic search engine   \cite{syntactic-search-2020}\footnote{\url{https://spike.apps.allenai.org}} which allows users to execute syntactic queries based on example sentences: the user enters a sentence satisfying the relation of interest and annotates it with light markup indicating the arguments and the trigger words. The system then automatically translates
 the markup into a syntactic pattern, matches it against a large pre-annotated corpus (\eg all Wikipedia sentences), and returns results. The user does not need to be familiar with syntactic formalisms or with advanced NLP.

\begin{table}[t]
\centering
\scalebox{0.95}{
\begin{tabular}{l|c|c}
{\textbf{Method}}               & \textbf{TACRED}  & \textbf{DocRED} \\ 
\hline
{Annotated 
{\small 5+50}}                & {0.097} & {0.140} \\ 
{Annotated {\small 10+100}}               & {0.136} & {0.215} \\ 
{Annotated {\small 20+200}}               & {0.266} & {0.271} \\ 
{Annotated {\small 50+500}}               & {0.458} & {0.311} \\ 
{Annotated {\small 100+1000}}              & {0.516} & {0.321} \\ 
{Annotated {\small  All}}              & {0.569} & {0.306} \\ 
\hline
{Pattern Based RE {\small (3 qrs)}}   & {0.128} & {-}     \\
\hline
{Synt. Search {\small (3 queries)}}       & {0.443} & {0.266} \\
{Example Generation}                   & {0.439} & {0.109} \\ 
{Search + Generation}    & {0.491} & {0.277} \\ 
\end{tabular}}
\caption{Average test F1 score over all relations. Pattern Based RE was given 3 positive patterns. Synt. Search is trained on data created from same 3 patterns. The Annotated experiments are denoted by the number of positive examples + negative examples.}
\label{tab:f1_table}
\end{table}

\subsection{By-example Patterns for Collecting Training Data}
Fig. \ref{fig:search-flow} demonstrates the user process. Starting with the sentence \emph{Paul founded Microsoft in April 1975}, the user marks \emph{Paul} as \emph{e1} (\texttt{e1:}) with an entity-type restriction of PERSON (\texttt{[e=PER]}), \emph{Microsoft} as \emph{e2} (\texttt{e2:}) with an entity-type ORG (\texttt{[e=ORG]}), and \emph{founded} as a trigger word (\texttt{\$founded}).
The SPIKE system translates the query into a syntactic graph, which is then matched against Wikipedia, returning 11,345 sentences matching the pattern (note that the word `founded' is matched lexically, while Paul and Microsoft become place holders for any person and any organization that adhere to the syntactic configuration). A subset of the returned sentences is then used as positive examples for model training.

While 11,345 cases make an impressive training set, these sentences share the same core syntactic configuration, and classifiers, trained on these matches, will not necessarily generalize well. The matches will also share the exact same lexical predicate (``founded''). The lack of lexical diversity of the predicate can be expanded by the user by supplying alternative words, perhaps aided by distributional similarity methods such as word2vec, or by querying a bi-LM such as BERT \cite{devlin2018bert} (\S\ref{exp:trigger_expansion}).
To counter the lack of structural diversity the user can supply additional patterns, derived from example sentences. For example, the user may supply also `[$_{e_2}$Microsoft]'s \underline{founder} [$_{e_1}$Paul]' (possessive construction) and `[$_{e_2}$Microsoft] was \underline{founded} by [$_{e_1}$Paul]' (passive) as additional patterns (\S\ref{sec:synt_search_results}).

\subsection{Experiments and Results}
\label{sec:synt_search_results}
\noindent\textbf{Setup}
For each relation, we select 3 representative sentences and annotate them based on the process described above\footnote{In this experiment, the selection of representative sentences is based on a heuristic process: we intuitively conceive of basic sentences exemplifying the relation, construct the corresponding Spike queries and briefly validate the number and quality of the returned results. We limit the number of seed examples to 3 since we believe coming up with 3 examples should be simple even for non-experts. In \S\ref{method:search-and-generation} we show that using more seed examples can further improve performance.}.
We do not perform any lexical expansion of trigger words beyond the initial pattern at this point.
The queries are processed by SPIKE~\cite{syntactic-search-2020} and the results are used as positive instances in the generated training set. A full list of the \spike~queries we used can be found in appendix \ref{appendix:patterns}, Table \ref{tab:positive_patterns}.

We also compare the TACRED classifier to a rule based extractor which uses the syntactic queries directly. Each syntactic query is added as a syntactic pattern to this extractor: any sentence which satisfies one of the syntactic patterns is labeled as a positive instance; sentences which do not satisfy any of the patterns are labeled negative. 

\noindent\textbf{Results}
Listed in the \emph{Synt. Search} and \emph{Pattern Based RE} rows of Table \ref{tab:f1_table}\footnote{Results correspond to 100+1,000 (\tacred) and 1,000+10,000 (\docred) examples, for results and discussions of different dataset sizes, see Appendix \ref{exp:more_syntactic}.},
\emph{Pattern Based RE}, using just the 3 patterns per relation, achieves a very low F-score of 12.8\%, due to low recall. However, this is already competitive with training a classifier on 5-10 positive examples per relation.
Training a classifier on the extracted relations increases the scores significantly, to 44.3$F_1$ on TACRED and 26.6$F_1$ on DocRED, approaching supervised training on 50+500 annotations (for TACRED) or 20+200 annotations (for DocRED).
 This result demonstrates that training an ML model over the output of a rule based model can significantly improve performance, echoing similar conclusions in \citet{angeli2013stanford}. Interestingly, \citet{angeli2013stanford} used a total of 4,697 patterns across 41 relations, an average of 114 patterns per relation. We demonstrate that by applying syntactic patterns to a large corpus and using modern DL classifiers, results competitive with manual annotation baselines can be reached with as few as 3 syntactic rules.

\begin{table}[tbp]
\centering
\begin{tabular}{llccc}
Dataset & Predicates & 100 & 500 & 1000 \\
\hline
\multirow{2}{*}{TACRED} & One Trig. & 0.487 & 0.459 & 0.461 \\
 & Trig. List & 0.517 & 0.490 & 0.478 \\
 \hline
\multirow{2}{*}{DocRED} & One Trig. & 0.290 & 0.336 & 0.338 \\
 & Trig. List & 0.316 & 0.338 & 0.337
\end{tabular}
\caption{F1 scores for \textit{founded by}, \textit{child}, \textit{place of death} and \textit{date of death} and \textit{spouse} when expanding the triggers list for the Syntactic Search ``by Example" method.}
\label{tab:triggers}
\end{table}

\subsubsection{Syntactic Search with Trigger Expansion}
\label{exp:trigger_expansion}

\noindent\textbf{Setup}
Constructing queries from 3 seed sentences produces retrieved sentences with low lexical diversity. \eg if all the seed sentences for \emph{founded-by} use the word \textit{``founded''} to express the relation, then all retrieved sentences will likewise include the word \textit{``founded''}, and exclude alternatives like \text{``established''}, \text{``formed''}, \text{``started''}, etc. 

In this experiment we generalize the seed queries to allow a list of trigger words rather than a single word. We consider only relations which include a lexical trigger in their seed patterns\footnote{\textit{per:children}, \textit{per:date of death}, \textit{org:founded by}, \textit{per:city of death} and \textit{per:spouse}, and \docred's \textit{child}, \textit{date of death}, \textit{founded by}, \textit{place of death} and \textit{spouse}}. Alternative triggers are selected by reviewing the closest words to the original triggers in word2vec's embedding space \cite{mikolov2013efficient}. Appendix \ref{appendix:predicate-expansion} includes the lists of alternative lexical triggers used. We train classifiers on 100+1000, 500+5,000 and 1,000+10,000 examples obtained from these expanded-trigger queries.

\noindent\textbf{Results}
As illustrated in Table \ref{tab:triggers}, adding alternative triggers improves results across all sample sizes for TACRED and for the 100+1000 size in DocRED.

\section{Augmenting Syntactic patterns with Natural Language Generation}

We showed how the \syntacticby~method works with only a few human annotated examples. In this section we would like to pursue NLG based methods to expand the number of exemplary patterns. Generative language models, compared to other methods for data augmentation (\eg Iterative bootstrapping and distant supervision) are highly accessible and require low technical expertise (sometimes passing a prompt is enough). Moreover, recent papers \cite{wei2019eda, anaby2019not, kumar2020data, amin2020exploring, russo2020control} report high impact of such models for the closely related Data Augmentation task. We therefore present numerous methods that take advantage of such models for RE bootstrapping.

First we show how a user can produce a high number of generated sentences using GPT2 \cite{radford2019language}. Then we demonstrate how the generated sentences can be integrated in the \syntacticby~method (\S\ref{method:search-and-generation}). Finally, in order to validate the necessity of the syntactic search in this flow we compare it to feeding the raw generations as inputs to a classifier (\S\ref{method:gen-to-classifier}).

\subsection*{Generating Examples LM}

\begin{figure}[t]
\centering
\includegraphics[width=\linewidth]{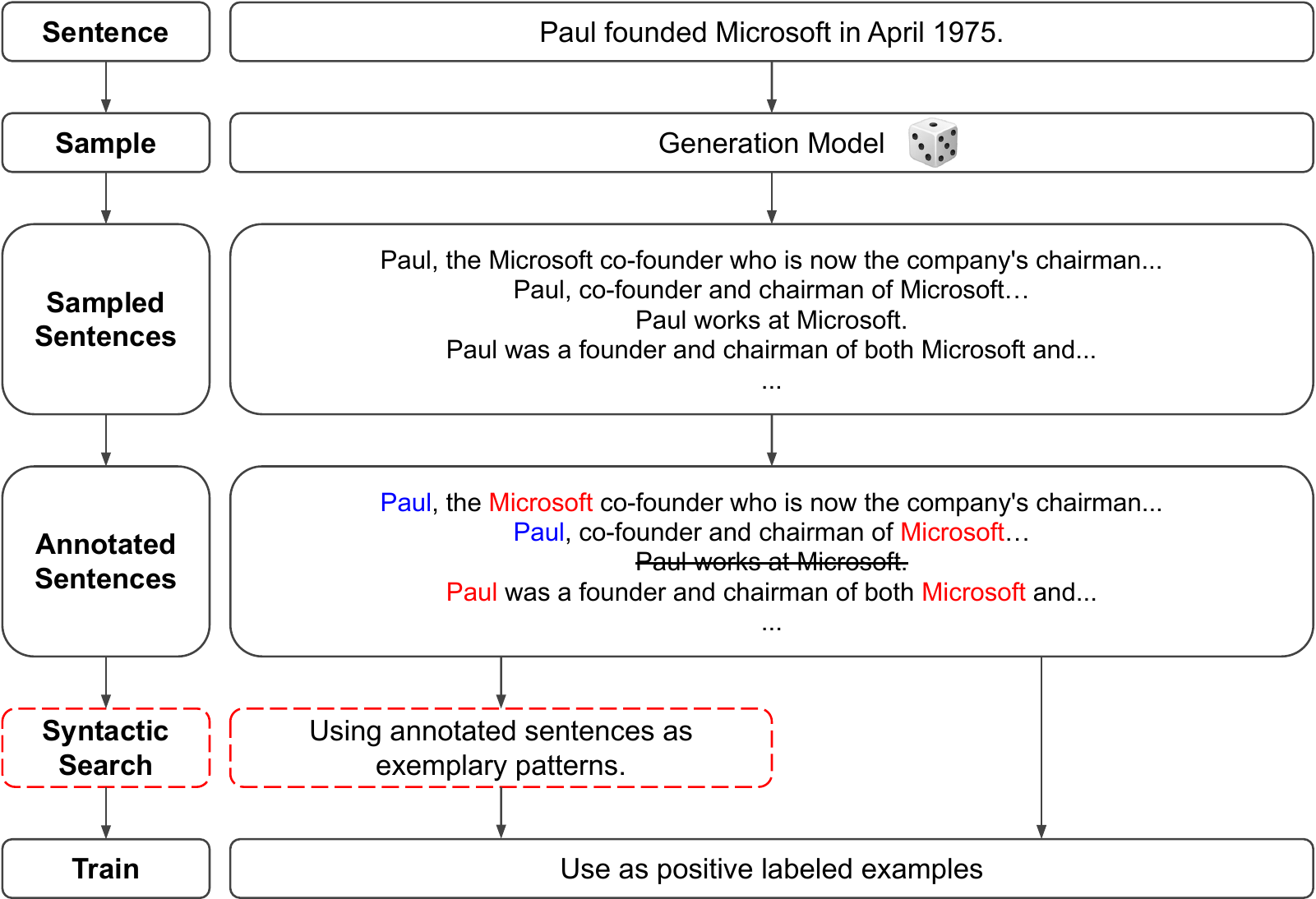}
\caption{Flow of sampling examples from conditional language model. The ``Syntactic Search" step corresponds to \S\ref{method:search-and-generation}, while skipping this step, corresponds to \S\ref{method:gen-to-classifier}. }
\label{fig:gen-flow}
\end{figure}

\paragraph{The user-flow} Depicted in Fig. \ref{fig:gen-flow}: The user enters a relation prompt (``Paul founded Microsoft''), to which the system responds by returning sentences that express the same relation. While not all returned sentences express the relation, many of them do. To filter out out-of-relation sentences the user goes through the list until she identifies a predefined number of positive examples (Here we used 100 sentences). In our experiments we encountered 1 positive example for every 3 examples annotated. This 1-out-of-3 ratio is significantly better than blindly sampling from a corpus (1-out-of-57, see \S\ref{data-annotation-challenge}), and by that can considerably save annotation time. For each example, the user marks the relevant entities, and optionally also the trigger word (the main word indicating the relation). These examples are then used as additional input examples to the syntactic search engine (\S\ref{method:search-and-generation}) or as train datasets for ML models (\S\ref{method:gen-to-classifier}).

\paragraph{Technical details}
We begin with a large pre-trained LM (we use GPT2-medium \cite{radford2019language}), and fine-tune it to the generation task. The method assumes the availability of relation-annotated data, though its relations do not need to overlap with the ones we are attempting to extract (in our case, we ensured the groups are distinct). The approach can be considered as an instance of transfer-learning, where we attempt to transfer the example-generation knowledge from the training relations to novel relations.
Given the annotated RE dataset, we consider positive examples of the form $(\bm{s}, e_1, e_2, r)$, where $r \in R$. We transform each instance to a conditioned LM training example, in which the LM sees a prefix (prompt) and should complete it. In our case the prompt is derived from $(e_1, e_2, r)$, followed by a special symbol, and we train the LM to produce the corresponding sentence $\bm{s}$. To derive the prefix we apply a pre-defined \emph{template} associated with each relation $r$\footnote{In our experiments, we use on average 3 different templates for each relation type, so a single annotated relation example will result in 3 (on average) different fine-tuning examples for the LM, each with a different prompt.}. The template has two slots to be filled with the entities $e_1$ and $e_2$. For example, a template for the \emph{founded-by} relation can take the form $[e_2]$ \emph{founded} $[e_1]$. We then fine-tune GPT2 on these training examples. 
At inference time, the user provides a \textit{single} prompt based on their desired relation.

Given the user prompt, we generate 1000 sentences with nucleus sampling \cite{holtzman2019curious} of $0.99$ and length of up to 50 tokens. We annotate the generated sentences until reaching 100 positive instances (usually requiring 200-300 sentences), this takes up to 1.5 hours per relation. These generated sentences are annotated and used as inputs to the syntactic search method (\S\ref{method:search-and-generation})  or directly as positive examples to a classifier (\S\ref{method:gen-to-classifier}).

\subsection{Enhancing Syntactic Search with Example Generation}
\label{method:search-and-generation}

We integrate the generation outputs in the \syntacticby~method by taking the positive annotated examples (on which we mark the entities as part of the annotation process) and automatically transforming them into SPIKE queries. This step has the potential to add substantial syntactic and lexical diversity to the pattern set, resulting in both larger and more diverse sets of positive examples. This combines the best of both worlds: the generative model is used to provide structural and lexical diversity, while the syntactic search system is used to provide a large selection of naturally occurring corpus sentences adhering to these patterns.

\subsubsection*{Experiments and Results}
\noindent
\noindent\textbf{Setup}
To reduce noise, we exclude queries where more than 1 out of 5 sampled results does not express the relation of interest. On average, we increased the number of syntactic patterns to 9.25, ranging from 6 to 14 after filtering.

\noindent\textbf{Results}
As listed in the \emph{Search + Generation} row of Table \ref{tab:f1_table}, this method achieved best performance for both \tacred~and \docred~with overall scores corresponding to 550/1100 and 220/550 annotated examples respectively. Using the generation outputs as examples doesn't only help in suggesting more sentences satisfying the relation but also in augmenting the number of predicates used. We looked on the number of predicates used for the \tacred~relations which include lexical triggers (similarly to \S\ref{exp:trigger_expansion}), the generation phase suggested $7.4$ predicates on average, more than the $2.8$ predicates per relation of our original patterns, and less than the trigger expansion method we suggested in \S\ref{exp:trigger_expansion}, where we tried to find all the possible predicates, with $18.2$ triggers on average. We conclude that while the \syntacticby~method performs well with only a few example patterns, this can be even improved with more input examples. While we report \syntacticby~enjoys such generation-based pattern augmentation, a similar boost with different, non-NLG, methods is of course possible. We leave further probing for other pattern augmentation methods as future work.

\subsection{Directly Training Classifiers using Generation Outputs}
\label{method:gen-to-classifier}

It is possible that generative models produce diverse enough training examples that will suggest our syntactic search superfluous.
We validate the necessity of taking the annotated generations (Annotated similarly to \S\ref{method:search-and-generation}) through the \syntacticby~method, by comparing it to simply passing the annotated generations as classifier inputs, as depicted in the RHS of Fig. \ref{fig:gen-flow}.

\subsubsection*{Experiments and Results}
\noindent
\noindent\textbf{Setup} Many of the samples include the entities from the prompt verbatim. Before using them as the model inputs, we replace the entities with a random Wikipedia entity of the same type.

\noindent\textbf{Results}
As can be seen in the \emph{Example Generation} row in Table \ref{tab:f1_table}, on \tacred, this method produces F1 scores on par with \syntacticby.  However, evaluating on \docred, the method does not produce competitive results\footnote{The language model used to generate examples was fine-tuned on a version of TACRED which excludes the relations we evaluate on. Still, for TACRED, the language model is fine-tuned and evaluated on data from the same domain (newswire). The DocRED data on the other hand, is taken from Wikipedia, so the evaluation is essentially out of domain. We therefore conclude that used independently, this approach is applicable only in cases where a background RE dataset is of the same domain as the target corpus from which we want to extract relations.}. On both datasets it produce worse than \textit{Search + Generation}. 
We conclude that it is more beneficial to use outputs of generative models as syntactic search queries, and by that find syntactically similar sentences, comparing to simply use generations as the train set. We deduce models are likely to generalize better on ``real world" examples.

\section{Additional Experiments}

\subsection{Results across relations}
\begin{figure}[t]
\centering
\includegraphics[width=\linewidth]{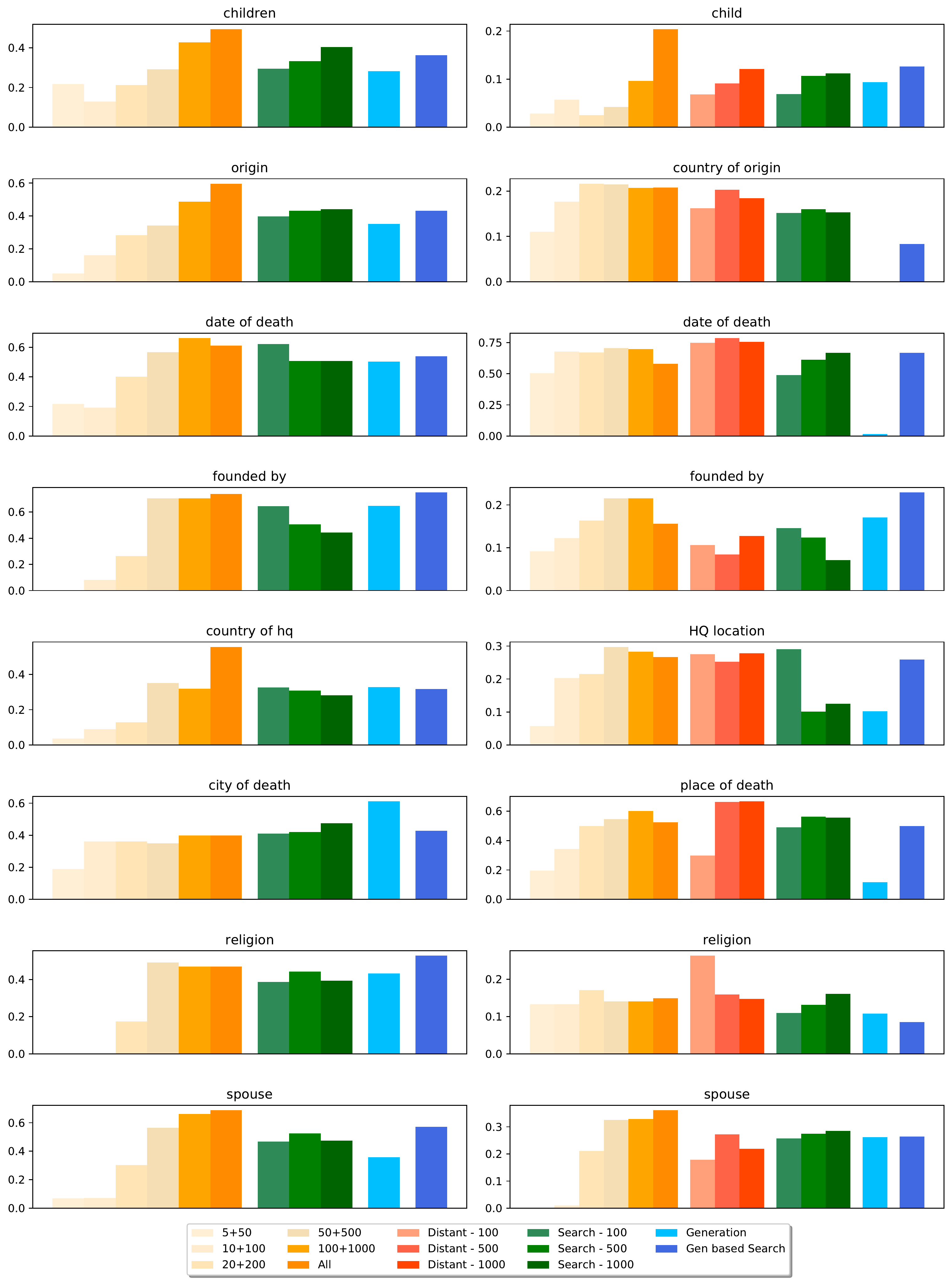}
\caption{F1 scores of \tacred~(right), and \docred~(left) by relation.}
\label{fig:tacred-and-docred}
\end{figure}

Analyzing the results we highlight some interesting trends (Fig. \ref{fig:tacred-and-docred}). First, we note that
the behavior is not consistent between relations, nor datasets: different relations behave differently, showing different trade-offs between different methods.

Classifiers for relations like \textit{``Religion"}\footnote{\tacred's ``religion" relation plateaus as it has a low number of train instances.}, \textit{``City of Death"} and \textit{``Date of Death"} seem to plateau at around 50-100 manually annotated examples. For these relations, annotating more data is not necessarily useful. The syntactic search approach works especially well for these relations: applying syntactic search over 3 seed queries is sufficient to yield results on par or slightly higher than all available manually annotated data. We hypothesize that these findings might be the result of low diversity in the ways these relations are typically expressed. 

While the combined \emph{Search + Generation} approach is overall useful, the effect is not consistent across relations: performance improves for some relations and deteriorates for others. In \S\ref{method:search-and-generation} we described the techniques we use to reduce the noise coming from additional queries. These techniques however are rather basic and these results indicate that more advanced techniques of the type used in \citet{angeli2015bootstrapped} and \citet{snorkel}, are likely to yield more consistent improvements.

\subsection{\distant}
\noindent\textbf{Setup}
Distant supervision \cite{mintz2009distant} suggests a method to construct a training dataset based on a large external KB of relation triplets. \citet{yao2019docred} offered a machine annotated version of \docred~constructed by aligning Wikipedia pages with Wikidata. The authors took great care in creating this resource: a high-quality NER model trained on in-domain manually annotated data was used to automatically annotate possible relation arguments; a named entity linker was used to merge entities with similar KB ID; and finally, Wikidata was queried in order to label pairs of linked entities. 

We trained a classifier using the released data, sampling increasing number of examples: (100+1,000, 500+5,000, 1,000+10,000). We report best score of 0.312$F_1$ (500+5,000 split).

\noindent\textbf{Results} This \distant~dataset, created by \citet{yao2019docred}, appears to be of very high quality and the results are on par with the full set of manually annotated data. These results indicate that given a large KB of relation triplets, a high-quality in-domain NER, and a high quality linking solution, distant-supervision is a very promising technique. It should be noted however, that the availability of all these external resources is very rare in practice and is not required by the methods proposed in this work.

\section{Applicability to other languages}
We explored only English in this work. However, we argue that our main method -- example-based syntactic search followed by DL-training -- is not strongly tied to English, and we encourage other researchers to experiment with it in their languages of interest. We provide details of what is needed to adapt the system to a different language. 

The \syntacticby ~method requires (1) An automatically dependency-parsed corpora in the language. These can be readily produced by the many syntactic parsers that are available for many languages \cite{manning2014stanford, spacy, qi2020stanza}.
(2) An indexing engine that supports efficient queries over parse trees. \citet{syntactic-search-2020} uses the open-source Odinson engine \cite{valenzuela2015domain} for this purpose.
(3) A component that translates a query in spike’s ``by example" syntax to the indexing engine’s query syntax. 
This requires finding the minimal (in terms of number of nodes) sub-graph that connects all relation arguments (and predicates if available), then search for sentences with similar sub-graphs in the index. With these three components, a syntactic-search system can be readily implemented. The rest of the components are straightforward application of DL methods. Indeed, we suspect the major obstacle in application to a new language will be the availability of evaluation data.

\section{Conclusion}
We show that with modern DL classifiers and a dataset bootstrapped using syntactic search with as few as 3 seed patterns can be as effective as a dataset with hundreds of manually annotated samples. Using LMs help to further diversify the dataset and improve results.
Overall, our results are positively optimistic for bootstrapping methods. However, this work is only an initial step in exploring methods for bootstrapping relation extractors using minimal user effort, supported by strong pre-trained neural LMs. We hope to encourage further work in this direction.

\section{Acknowledgments}
This project has received funding from the European Research Council (ERC) under the European Union’s Horizon 2020 research and innovation programme, grant agreement No. 802774 (iEXTRACT).

\bibliographystyle{acl_natbib}
\bibliography{eacl2021}

\appendix

\section{Estimated effort for Annotation}
\label{appendix:effort}
Table \ref{table:hg_means} lists for each relation the ratio of positive to negative examples in the TACRED training set. A negative example for a relation \emph{r}, is any example whose entities share types with a positive example of \emph{r}, but whose label is different from \emph{r}. Note that TACRED  significantly under represents negative examples so the reported ratio is an upper bound on the ratio in the wild.

\begin{table}[htbp]
\centering
\begin{tabular}{l|c}
Relation                                      & Pos/Neg Ratio \\
\hline
org:country\_of\_hq                           & 1/7 \\
org:founded\_by                               & 1/56 \\
per:children                                  & 1/64 \\
per:city\_of\_death                           & 1/31 \\
per:date\_of\_death                           & 1/26 \\
per:origin                                    & 1/10  \\
per:religion                                  & 1/2 \\
per:spouse                                    & 1/52
\end{tabular}
\caption{Pos/Neg ratio in TACRED, rounded to the closest fraction.}
\label{table:hg_means}
\end{table}



\section{\syntacticby~with varying dataset sizes}
\label{exp:more_syntactic}
We experimented with varying the number of sampled examples, using the same 3 seed syntactic patterns. The results are reported in Table \ref{tab:size_search}. While \docred's F1 scores increase with increasing number of sampled examples, the trend is opposite in \tacred. We believe this is due to different initializations and inductive noise in both the positive and negative samples introduced by sampling from semi-noisy data.

\begin{table}[htbp]
\centering
\begin{tabular}{l|c|c}
{\textbf{Method}}       & \textbf{TACRED} & \textbf{DocRED} \\ 
\hline
{Synt. Search - 100}       & {0.443} & {0.250} \\ 
{Synt. Search - 500}       & {0.434} & {0.259} \\ 
{Synt. Search - 1000}      & {0.427} & {0.266} \\ 
\end{tabular}
\caption{\syntacticby~with different training sizes}
\label{tab:size_search}
\end{table}

\section{Trigger List Expansion}
\label{appendix:predicate-expansion}
For the majority of patterns used in the \syntacticby~experiments we used a single trigger word (see Appendix \ref{appendix:patterns}). To experiment with using trigger lists, we modified the patterns in Appendix \ref{appendix:patterns} in the following way:

We changed the triggers in all 
\emph{child\textbackslash children}
patterns to include any of the following possibilities: 
\begin{description}
    \item baby, child, children, daughter, daughters, son, sons, step-daughter, step-son, step-child, step-children, stepchildren, stepdaughter, stepson
\end{description}

For \emph{founded-by} relations we change the ``founder" trigger to be any of these triggers:
\begin{description}
    \item founder, co-founder, cofounder, creator
\end{description}
and changed ``founded" to be any trigger from the following list:
\begin{description}
    \item create, creates, created, creating, creation, co-founded, co-found, debut, emerge, emerges, emerged, emerging, establish, established, establishing, establishes, establishment, forge, forges, forged, forging, forms, formed, forming, founds, found, founded, founding, launched, launches, launching, opened, opens, opening, shapes, shaped, shaping, start, started, starting, starts
\end{description}

In \emph{spouse} relations we expanded the ``husband\textbackslash wife" trigger to be any of:
\begin{description}
\item ex-husband, ex-wife, husband, widow, widower, wife, sweetheart, bride
\end{description}
and the ``marry" trigger to:
\begin{description}
\item divorce, divorced, married, marry, wed, divorcing
\end{description}

For the \emph{``date of death"} and \emph{``place\textbackslash city of death"} we changed the ``died" trigger to any of:
\begin{description}
\item died, executed, killed, dies, perished, succumbed, passed, murdered, suicide
\end{description}

\onecolumn
\section{Examples used for Syntactic Search by Example}
\label{appendix:patterns}
\begin{longtable}{| p{\textwidth} |}
\endfirsthead
\hline
\endfirsthead
\endhead
\multicolumn{1}{r}{\textit{Continued on next page}} \\
\endfoot
\endlastfoot
\textbf{\textit{child}} \\
<>e1:[e=PER]John 's t:[w=\{triggers\}]daughter , <>e2:[e=PER]Tim, likes swimming. \\[0.5em]
<>e1:[e=PER]Mary did something to her t:[w=\{triggers\}]son, <>e2:[e=PER]John in 1992. \\[0.5em]
<>e1:[e=PER]Mary was survived by her 4 t:[w=\{triggers\}]sons, John, John, <>e2:[e=PER]John and John.\\[0.5em]
triggers =  son | daughter | child | children | daughters | sons
\\
\hline
\textbf{\textit{founded by}} \\
<>e1:[e=ORG]Microsoft t:[w]founder <>e2:[e=PER]Mary likes running. \\[0.5em]
<>e2:[e=PER]Mary t:[w]founded <>e1:[e=ORG]Microsoft. \\[0.5em]
<>e1:[e=ORG]Microsoft was t:[w]founded \$by <>e2:[e=PER]Mary. \\
\hline
\textbf{\textit{headquarters location}} \\
John Doe, a professor at the <>e1:[e=ORG]Oxford <>in:[t=IN]in <>e2:[e=LOC]England likes running. \\[0.5em]
<>e1:[e=ORG]Oxford, a leading <>t:[t=NN]company <>in:[t=IN]in <>e2:[e=LOC]England. \\[0.5em]
<>e2:[e=LOC]England pos:[t=POS]'s largest university is <>e1:[e=ORG]Oxford. \\
\hline
\textbf{\textit{religion}} \\
<>e1:[e=PER]John is a e2:[w=\{triggers\}]Jewish,, \\[0.8em]
e2:[w=\{triggers\}]Jewish <>e1:[e=PER]John is walking down the street. \\[0.8em]
<>e1:[e=PER]John is a e2:[w=\{triggers\}]Methodist Person.
\\[0.5em]
triggers =  Methodist | Episcopal | separatist | Jew | Christian | Sunni | evangelical | atheism | Islamic | secular | fundamentalist | Christianist | Jewish | Anglican | Catholic | orthodox | Scientology | Islamist | Islam | Muslim | Shia
\\
\hline
\textbf{\textit{spouse}} \\
<>e1:[e=PER]John 's t:[w=wife | husband]wife, <>e2:[e=PER]Mary , died in 1991. \\
<>e1:[e=PER]John t:[l]married <>e2:[e=PER]Mary,, \\
<>e1:[e=PER]John is t:[w]married to <>e2:[e=PER]Mary, \\
\hline
\textbf{\textit{origin}} \\
<>e2:[e=MISC]Scottish <>e1:[e=PER]Mary is high. \\
<>e1:[e=PER]Mary is a <>e2:[e=MISC]Scottish professor. \\
<>e1:[e=PER]Mary, the <>e2:[e=LOC]US professor. \\
\hline
\textbf{\textit{date of death}} \\
<>e1:[e=PER]John was announced t:[w]dead in <>e2:[e=DATE]1943. \\
<>e1:[e=PER]John t:[w]died in <>e2:[e=DATE]1943. \\
<>e1:[e=PER]John, an NLP scientist, t:[w]died <>e2:[e=DATE]1943. \\
\hline
\textbf{\textit{place of death}} \\
<>e1:[e=PER]John t:[w]died in <>e2:[e=LOC]London, <>country:e=LOC England in 1997. \\
<>e1:[e=PER]John t:[w]died in <>e2:[e=LOC]London in 1997. \\
<>e1:[e=PER]John \$-LRB- t:[w]died in <>e2:[e=LOC]London \$-RRB-. \\
\hline
\textbf{\docred's \textit{founded by}} \\
<>e1:[e=ORG]MISC Microsoft t:[w]founder <>e2:[e=PER]Mary likes running. \\
<>e2:[e=PER]Mary t:[w]founded <>e1:[e=ORG]MISC Microsoft. \\
<>e1:[e=ORG]MISC Microsoft was t:[w]founded \$by <>e2:[e=PER]Mary. \\
\hline
\textbf{\docred's \textit{origin}} \\
<>e2:[e=MISC]Scottish company, <>e1:[e=ORG]Microsoft is successful. \\
<>e1:[e=ORG]MISC Microsoft is a <>e2:[e=MISC]Scottish Company. \\
<>e1:[e=ORG]MISC Microsoft is a <>t:[t=NN]song \$by <>e2:[e=MISC]Scottish musician. \\
\hline
\textbf{\docred's \textit{date of death}} \\
<>e1:[e=PER]John \$-LRB- \\
<>e1:[e=PER]John t:[w]died in <>e2:[e=DATE]1943. \\
<>e1:[e=PER]John, an NLP scientist, t:[w]died <>e2:[e=DATE]1943. \\
\hline
\textbf{\docred's \textit{place of death}} \\
<>e1:[e=PER]John t:[w]died in <>e2:[e=LOC]London, <>country:e=LOC England in 1997. \\
<>e1:[e=PER]John t:[w]died in <>e2:[e=LOC]London in 1997. \\
<>e1:[e=PER]John \$-LRB- \$[e=DATE]1997, \$[e=LOC]London \$- \$[e=DATE]1997 <>e2:[e=LOC]London \$-RRB-. \\
\hline
\textbf{\docred's \textit{headquarters location}} \\
<>e1:[e=ORG]Microsoft, a leading <>t:[t=NN] company <>in:[t=IN]in <>e2:[e=LOC]Redmond. \\
<>e1:[e=ORG]Microsoft is t:[l=base | headquarter]based in <>e2:[e=LOC]England. \\
<>e1:[e=ORG]Microsoft, a leading <>t:[t=NN] company based <>in:[t=IN]in <>e2:[e=LOC]Redmond. 
\\
\hline
\caption{SPIKE search patterns for \tacred~and \docred~relations.}
\label{tab:positive_patterns}
\end{longtable}

\end{document}